%% file: new_draft.tex
\documentclass{article}
\usepackage{spconf,amsmath,graphicx}
\usepackage{array}
\usepackage{booktabs}
\usepackage{multirow}
\usepackage{amssymb}
\usepackage{xcolor}
\usepackage{physics}
\usepackage{subcaption}
\usepackage{url}

\title{Deeply Supervised Multimodal Attentional Translation Embeddings for Visual Relationship Detection}

\name{Nikolaos Gkanatsios$^{1,2}$,  Vassilis Pitsikalis$^1$,  Petros Koutras$^2$,  Athanasia Zlatintsi$^2$,  Petros Maragos$^2$}
\address{
	$^1$DeepLab, 11635, Athens, Greece\\ 
	$^2$School of E.C.E., National Technical University of Athens, 15773, Athens, Greece\\ 
	\small{Email: nikos.gkanatsios93@gmail.com, vpitsik@deeplab.ai,  \{pkoutras, nzlat, maragos\}@cs.ntua.gr}{\thanks{{This project was conducted while N. Gkanatsios was an intern at the National Technical University of Athens and has been funded by deeplab.ai, as far as V. Pitsikalis and N. Gkanatsios are concerned. This is a part of the deeplab.ai research activities, such as student research-training funding, and collaborations with academic institutions.} }}}
\begin{document}
	\ninept
	\maketitle
	\begin{abstract}
%		Detecting visual relationships, i.e. $<$Subject, Predicate, Object$>$ triplets, in an image is a challenging Scene Understanding task. Prior works have shown remarkable progress by taking advantage of linguistic priors or spatial information, however, most of them train a single feature path that performs either a feature-level or a score-level fusion of the different modalities. We introduce Multimodal Attentional Translation Embeddings (MATransE), a two-branch architecture consisting of a Predicate branch and an Object-Subject Branch, jointly optimized using Deep Supervision. The visual features of each branch are driven by a multimodal attentional mechanism that projects relationships into a low-dimensional space where spatio-linguistically close relationships come closer. Experiments on different metric settings on VRD dataset \cite{Lu_2016} show that MATransE outperforms other approaches %by at least XX\%,
%		bringing the task to a new state-of-the-art. Our code will be made publicly available.
		Detecting visual relationships, i.e. $<$Subject, Predicate, Object$>$ triplets, is a challenging Scene Understanding task approached in the past via linguistic priors or spatial information in a single feature branch. We introduce a new deeply supervised two-branch architecture, the Multimodal Attentional Translation Embeddings, where the visual features of each branch are driven by a multimodal attentional mechanism that exploits spatio-linguistic similarities in a low-dimensional space. We present a variety of experiments comparing against all related approaches in the literature, as well as by re-implementing and fine-tuning several of them. Results on the commonly employed VRD dataset \cite{Lu_2016} show that the proposed method clearly outperforms all others, while we also justify our claims both quantitatively and qualitatively.
	\end{abstract} 
	\begin{keywords}
		Visual Relationship Detection, Attention, Multiple Modalities, Translation Embeddings, Deep Supervision.
	\end{keywords}
	
	\section{Introduction}
	\input{tex_files/section_1.tex}

	\begin{figure*}[!t]
		\centering
		\includegraphics[width=.8\textwidth]{figs/matranse2.png}%[width=2.9in,height=1.6in]s
		\caption{Overview of the proposed approach. Given a detected subject-object pair, MATransE employs spatial and linguistic information (binary masks and word embeddings) in an attention module (SLA-M) that guides the classification of the convolutional features of the subject (orange), object (green) and predicate box (purple). Subject-object and predicate features are handled into two separate branches, OS-Branch and P-Branch respectively, and their scores are fused. The single branches and their fusion all contribute to the total loss (terms $\mathbf{\mathcal{L}_{OS}}, \mathbf{\mathcal{L}_{P}}, \mathbf{\mathcal{L}_{f}}$) and are jointly optimized with Deep Supervision, forcing an alignment in the score space such that $P \approx O - S$.}
		\label{system}
	\end{figure*}

	\section{Related Work}

\input{tex_files/section_2.tex}

	\section{Methodology}
	
	\input{tex_files/section_3.tex}

	\section{Evaluation and Results} \label{sec4}
	
	\input{tex_files/section_4.tex}

	\section{Conclusion}
	\input{tex_files/section_5.tex}
	
	\bibliographystyle{IEEEbib}
	\newpage
	\bibliography{Bibliography}
	
\end{document}

%% file: tex_files/section_1.tex
\begin{figure}[!t]
		\centering
		\includegraphics[width=.46\textwidth]{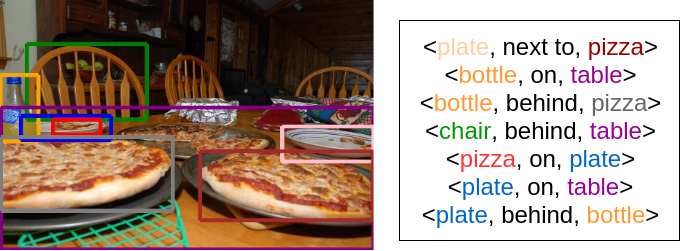}%s
		\caption{Visual Relationship Detection refers to localizing pairs of objects ($S$, $P$) and classifying their interactions $P$ to form $<S, P, O>$ triplets. The exponential number of possible such relationships and the visual variability of $P$ -- e.g. ``behind" on ``bottle - behind - pizza'' and ``chair - behind - table'' -- render the task quite challenging.}
		\label{VRD_examples}
	\end{figure}

%In this work we aim to classify semantic visual relationships between objects such as: 
Is this ``pizza slice \emph{on} the plate" or ``\emph{next to} the plate?" The classification of such semantic visual relationships, namely Visual Relationship Detection, aims on bridging the gap between visual and semantic perception by detecting $<S, P, O>$ triplets, with the predicate $P$ being the relationship of subject $S$ and object $O$, e.g. ``pizza - on - plate" (Fig.~\ref{VRD_examples}). The large intra-class variance of the predicates' appearance and the vast number of unbalanced classes led prior works to learn $S$, $P$ and $O$ separately and employ linguistic \cite{Lu_2016,Zhuang_2017} and spatial information \cite{Zhu_2017,Liang_2018} to deal with the predicates' visual variance. Recently, \cite{Zhang_2017} proposed Visual Translation Embeddings (VTransE), a representation of relationships as a vector translation, e.g. $\mathrm{cat} + \mathrm{eat} \approx \mathrm{fish}$. Albeit competent, VTransE uses only subject and object features and no predicate or contextual information, proved effective lately \cite{Zhuang_2017,Yin_2018}.
	
%	Inspired by Visual Translation Embeddings \cite{Zhang_2017}, the complementary nature of language and vision \cite{Lu_2016} and recent advances in Attention mechanisms \cite{Jetley_2018} and Deep Supervision \cite{Xie_2015}, we introduce Multimodal Attentional Translation Embeddings (MATransE) that 1) directly use predicate features as well as subject and object features, in two separate branches, 2) employ a spatio-linguistic attention mechanism to drive the visual features, 3) perform an alignment of the two branches in a common score space using Deep Supervision. We evaluate our system on the widely used VRD \cite{Lu_2016} dataset, shedding light to inconsistencies in the metric's definition that made prior works incomparable and re-implement several baselines for a fair comparison. Our system outperforms other methods, proving the superiority of our approach against prior work.
	
	Inspired by VTransE \cite{Zhang_2017}, the complementary nature of language and vision \cite{Lu_2016} and the recent advances in attention mechanisms \cite{Jetley_2018} and Deep Supervision \cite{Xie_2015}, we introduce Multimodal Attentional Translation Embeddings (MATransE): 1)~We directly use predicate features and subject-object features, in two separate branches, 2)~we employ a spatio-linguistic attention mechanism to drive the visual features, 3)~we perform an alignment of the two branches in a common score space using Deep Supervision. Our system is evaluated on the widely used VRD \cite{Lu_2016} dataset. We also shed light to inconsistencies in the metric's definition that made prior works incomparable and re-implement and fine-tune several baselines so as to be comparable. Our system outperforms all other methods, showing the importance of the proposed approach.

%% file: tex_files/section_2.tex
\textit{Visual Relationships Detection}~\cite{Sadeghi_2011} was only recently formulated \cite{Lu_2016}. Most works detect objects \cite{Ren_2015} and then predict the predicate class of each object pair \cite{Lu_2016,Zhuang_2017,Zhu_2017,Liang_2018}. % and some refine subject-object scores \cite{Dai_2017,Zhang_2017}. 
We cluster the related literature based on the feature type and the fusion level of different modalities.

\textit{Visual Appearance Features} are extracted from the predicate box, i.e. the minimum rectangle that encompasses the subject box and the object box \cite{Lu_2016,Cui_2018,Zhuang_2017,Goutsu_2018,Yu_2017,Zhu_2017}, the separate subject-object boxes \cite{Zhang_2017,Plesse_2018,Yang_2018,Peyre_2017}, or both \cite{Baier_2017,Zhu_2018,Liang_2018,Zhang_2018,Zhang_2018b}. All the above train a single branch with visual features, while we jointly train two separate branches with different features, a predicate feature branch (P-branch) and an object-subject branch (OS-branch), and employ Deep Supervision to align their scores into a common space.

\textit{Linguistic and Semantic Features} are  employed in a feature-level integration with word embeddings \cite{Lu_2016,Goutsu_2018,Liang_2018,Cui_2018,Zhu_2017}, encoding of statistics \cite{Baier_2017,Goutsu_2018,Liang_2018,Yu_2017}, late-fusion with subject-object classemes (score vectors) \cite{Dai_2017,Zhang_2017,Yu_2017,Zhu_2018} and loss-level fusion as regularization terms \cite{Lu_2016,Zhu_2017} or adaptive-margins \cite{Liang_2018,Zhu_2018,Cui_2018}. Closest to us, \cite{Zhuang_2017} uses subject-object embeddings to train context-aware classifiers and \cite{Zhang_2018} trains multimodal embeddings by projecting visual and linguistic features into a common space. Different from these approaches, we use language as a component of our multimodal attentional scheme to drive the visual features of the two branches.

Lastly, \textit{Spatial Features}, either hand-crafted \cite{Zhu_2017,Goutsu_2018,Yu_2017,Peyre_2017} or convolutional \cite{Dai_2017,Liang_2018,Zhu_2018,Cui_2018}, are often integrated in a feature- \cite{Dai_2017,Liang_2018,Zhu_2018,Cui_2018}, score- \cite{Zhang_2018b} or loss-level fusion \cite{Zhu_2017}. Recently, \cite{Kolesnikov_2018} used masks as an early-stage attention mechanism to detect objects related to a given one. We integrate mask features into our spatio-linguistic attention scheme to control visual features' classification.

\textit{Visual Translation Embeddings} \cite{Zhang_2017} is the most related work, mapping $S$, $P$ and $O$ in a vector space where valid relationships satisfy $S + P \approx O$. Nonetheless, no predicate features are used, leaving the alignment of the two terms of the equation to take place inside the loss function. Instead, our architecture is two-branch and its components are jointly optimized using Deep Supervision. Predicate and subject-object features are explicitly used, guided by a spatio-linguistic attention mechanism.

%% file: tex_files/section_3.tex
\subsection{Background}
	Visual Relationship Detection involves simultaneous detection of all $<S, P, O>$ triplets in a given image by localizing $S$ and $O$ and classifying their interaction $P$. In addition to the high visual variability of $S$, $P$, $O$, another challenge is the large number of possible triplets, e.g. 100 object and 70 predicate classes yield 700000 relationships. We adopt an approach scalable to thousands of relationships by learning to predict each component separately \cite{Lu_2016}, factorizing the Visual Relationship Detection process as:
	\begin{equation}\label{eqfactorize}
	Pr(<S, P, O>|I) = Pr(S,O|I) Pr(P|S,O,I),
	\end{equation}
	separating object detection (1st term) from predicate classification (2nd term). The first can be addressed by a state-of-the-art object detector like \cite{Ren_2015,He_2017}, so the rest of this paper focuses on the more challenging second one.  For this term, \cite{Zhang_2017} exploits features from the detected subject and object and models relationships by projecting $S$, $P$ and $O$ into an embedding space where:
	\begin{equation}\label{eqVTransE}
	S + P \approx O
	\end{equation}
	Following this formulation, we introduce Multimodal Attentional Translation Embeddings, MATransE, that guides the features' projection with attention and Deep Supervision to satisfy Eq.~\ref{eqVTransE}.
	
	\subsection{Multimodal Attentional Translation Embeddings}
	MATransE learns a projection of $<S, P, O>$ into a score space where $S + P \approx O$, by jointly representing $S$, $P$ and $O$ as $\mathbf{W}_S x_S$, $\mathbf{W}_P x_P$ and $\mathbf{W}_O x_O$. Vectors $x_S, x_P, x_O$ are visual appearance features \cite{He_2015}, extracted both from the predicate image (union of subject and object bounding boxes) and the subject and object images of a pair of detected objects. To learn the projection matrices $\mathbf{W}_S, \mathbf{W}_P, \mathbf{W}_O$, MATransE employs a Spatio-Linguistic Attention module (SLA-M) that uses binary masks' convolutional features $m$ \cite{Dai_2017} and encodes subject and object classes with pre-trained word embeddings $s, o$ \cite{Mikolov_2013, Lu_2016}. Therefore, Eq.~\ref{eqVTransE} becomes:
	\begin{equation}\label{eqMATransE}
	\mathbf{W}_P(s, o, m) x_P = \mathbf{W}_O(s, o, m) x_O - \mathbf{W}_S(s, o, m) x_S
	\end{equation}
	This allows us to design a two-branch architecture: a branch driving the predicate features into scores $\mathbf{W}_P(s, o, m) x_P$ (P-branch) and another classifying the difference of weighted object-subject features $\mathbf{W}_O(s, o, m) x_O - \mathbf{W}_S(s, o, m) x_S$ (OS-branch). To satisfy Eq.~\ref{eqMATransE}, we enforce a score-level alignment by jointly minimizing the loss of each one of the P- and OS-branch with respect to ground-truth using Deep Supervision \cite{Xie_2015}. See Fig.~\ref{system} for an overview.
	
	\textit{Spatio-Linguistic Attention Module (SLA-M)} combines a 3-layer CNN \cite{Dai_2017} to encode binary mask features and a MLP to encode subject and object embeddings. The spatial and linguistic information are fused with a fully-connected (FC) layer that performs a projection of the spatio-linguistic vector into a low-dimensional (64-d) space where spatially and semantically similar configurations come closer. %while either spatially or semantically different configurations lie further.
	 For instance, ``man on horse" might be spatio-linguistically closer to ``man ride elephant" than ``man next to horse" and ``cat on horse" that share either the same linguistic, i.e. ``man'', ``horse'', or spatial, i.e. ``on'', information.
	
	\textit{Predicate Branch}: Predicate images often contain irrelevant information that ``distracts" the network, thus, we use pre-pooled predicate features that have spatial dimensions as well and apply attentional pooling \cite{Zhuang_2017} to concentrate the network's ``focus" on the discriminative visual cues that may only appear in a small fraction of the image. Intuitively, the attention function should be different depending on the objects' classes and their spatial configuration. Thus, our attentional pooling weights $\mathbf{W}_{att}^{(P)}(s, o, m)$ are directly dependent on SLA-M's output.
	
	The pooled feature vector is further encoded via a FC layer and is classified into predicate scores using an attentional classifier. Once again, it is easier to classify the predicate knowing that the relationship concerns certain objects at specific relative position and sizes. We thus obtain the P-branch scores as:
	\begin{equation}\label{eqPScores}
	P = \mathbf{f}_P \mathbf{W}_{cls}^{(P)}(s, o, m) + \mathbf{b}_{cls}^{(P)}
	\end{equation}
	with $\mathbf{f}_P$ the encoded predicate feature vector, $\mathbf{W}_{cls}^{(P)}(s, o, m)$ the spatio-linguistic classifier weights and $\mathbf{b}_{cls}^{(P)}$ a bias vector (Fig.~\ref{system}).
	
	\textit{Object-Subject Branch}: Subject and object regions are far less noisy and we therefore consider pooled feature vectors to be a meaningful representation. Contrary to \cite{Zhang_2017}, we do not subtract the subject and object features directly but we first encode them as $\mathbf{f}_S$, $\mathbf{f}_O$ using two FC layers, as the subject and object regions often occlude each other, so that pure subtraction would possibly weaken meaningful features. The difference of the encoded features is classified using an attentional classifier, similar to the P-branch:
	\begin{equation}\label{eqOSScores}
	O - S = (\mathbf{f}_O - \mathbf{f}_S) \mathbf{W}_{cls}^{(O-S)}(s, o, m) + b_{cls}^{(O-S)}
	\end{equation}
	with $\mathbf{W}_{cls}^{(O-S)}(s, o, m)$ denoting the spatio-linguistic attentional classifier weights and $b_{cls}^{(O-S)}$ a bias vector.
	
	\textit{Fusion with Deep Supervision}: We combine P- and OS- branches' scores into a single vector and train a meta-classifier to obtain the predicate classes, minimizing the fusion loss $\mathcal{L}_f$. The scores of each branch are deeply supervised, i.e. we demand that they give a meaningful prediction as well, minimizing $\mathcal{L}_P$ and $\mathcal{L}_{OS}$ respectively. Thus, with $\mathbf{W} = (\mathbf{W}_P, \mathbf{W}_O, \mathbf{W}_S)$, the total loss:
	\begin{equation}\label{eqTotalLoss}
	\mathcal{L}(\mathbf{W}) = \lambda_f \mathcal{L}_{f}(\mathbf{W}) + \lambda_P \mathcal{L}_{P}(\mathbf{W}_P) + \lambda_{OS} \mathcal{L}_{OS}(\mathbf{W}_O, \mathbf{W}_S)
	\end{equation}
	where $\lambda$ hyperparameters balance each term's importance.

%% file: tex_files/section_4.tex
\subsection{Dataset and Metrics} \label{sec4.1}
	We evaluate our approach on the widely benchmarked VRD \cite{Lu_2016}; this contains 5000 images with 100 object categories, 70 predicates, 37993 annotated relationships and 6672 relationship types.
	We report results for Predicate Classification, a task that directly quantifies Visual Relationship Detection. In this setup, the objects' boxes and classes are given and the task is to predict the predicate.
	
	As explained in \cite{Lu_2016}, a suitable metric is $\mathrm{Recall}@x$, that counts the fraction of times the correct relationship is included in the top $x$ confident predictions, where $x=50, 100$. However, there is one hyperparameter $k$ in the metric computation that prior works often do not specify, leading to unfair comparisons. To address this, we re-formulate the metric as $\mathrm{Recall}_k@x$ ($R_k@x$). Let $N$ be the number of examined subject-object pairs in an image; then, keeping the top-$k$ predictions per pair, $R_k@x$ examines the $x$ most confident predictions out of $Nk$ total.% If we choose $k=1$, our definition is consistent with \cite{Lu_2016}. 
	
	Most works have seen Predicate Classification as a multiclass problem and they use $k=1$ to reward the correct top-1 prediction for each pair \cite{Lu_2016,Peyre_2017,Zhu_2017,Zhang_2017,Zhuang_2017}. Motivated by the fact that there are pairs annotated with more than one predicate classes, other works \cite{Liang_2018,Cui_2018} have tackled this as a multilabel problem and they use $k=70$ to allow for predicate co-occurrences \cite{Liang_2018,Cui_2018}. Intuitively, the two choices do not always converge to the same optimum, as the one favors only the most probable class, while the other does not force any of the true labels to be the top prediction, leaving prior works incomparable due to their different viewpoint of the task itself.

	\begin{table}
		\centering
		\footnotesize
		\begin{tabular}{c|c|c|c}
			\hline
			\textbf{Method} & $\mathbf{R_1@50}$ & $\mathbf{R_{70}@50}$ & $\mathbf{R_{70}@100}$ \\ \hline
			
			VTransE \cite{Zhang_2017} & 44.7 & - & - \\ \hline
			
			PPR-FCN \cite{Zhang_2017b} & 47.7 & - & - \\ \hline
			
			VRD \cite{Lu_2016} & 47.8 & - & - \\ \hline
			
			STA \cite{Yang_2018} & 48.03 & - & - \\ \hline
			
			S-R \cite{Zhu_2017} & 51.5 & - & - \\ \hline
			
			Weak Supervision \cite{Peyre_2017} & 52.6 & - & - \\ \hline
			
			Sem. Modeling \cite{Baier_2017} & 53.1 & - & - \\ \hline
			
			CAI \cite{Zhuang_2017} & 53.59 & - & - \\ \hline
			
			Guided Proposals \cite{Plesse_2018} & - & 71.3 & 81.8 \\ \hline
			
			DR-Net \cite{Dai_2017} & - & 80.7 & 81.9 \\ \hline
			
			ROR \cite{Goutsu_2018} & - & - & 82.1 \\ \hline
			
			Bi-RNN \cite{Liao_2017} & - & 84.9 & 92.6 \\ \hline
			
			DSR \cite{Liang_2018} & - & 86.0 & 93.1 \\ \hline
			
			CDDN \cite{Cui_2018} & - & 87.57 & 93.76 \\ \hline
			
			Zoom-Net \cite{Yin_2018} & 50.6 & 84.2 & 90.5 \\ \hline
			
			LK (VRD+VG) \cite{Yu_2017} & 54.82 & 83.97 & 90.63 \\ \hline
			
			Zoom-Net + CAI \cite{Yin_2018} & \underline{55.9} & \underline{89.0} & \underline{94.5} \\ \hline \hline
			
			VTransE (ours) & 46.2 & 78.2 & 87.8 \\ \hline
			
			VRD (ours) & 47.8 & 81.7 & 91.1\\ \hline
			
			CAI (ours) & 51.0 & 85.69 & 93.8 \\ \hline
			
			ROR (ours) & 53.0 & 82.8 & 92.1 \\ \hline
			
			DR-Net (ours) & 53.79 & 79.47 & 91.31 \\ \hline \hline
			
			\textbf{Ours MATransE} & \textbf{56.14} & \textbf{89.79} & \textbf{96.26} \\ \hline
		\end{tabular} 
		\caption{Comparisons on VRD Predicate Classification with all related reported and re-implemented (``(ours)") results. Bold fonts indicate best performance and underlined ones the 2nd best. Note that $\mathbf{R_1@50} = \mathbf{R_1@100}$ \cite{Lu_2016} on VRD, thus we report only $\mathbf{R_1@50}$.}% ``(ours)" denotes our re-implemention.}
		\label{tab:Results}
	\end{table}

	\subsection{Experimental results}
%	We implement our full system in PyTorch, based on the ResNet-101 \cite{He_2015} for feature extraction\footnote{We train our model for 10 epochs using batch size 32 and Adam optimizer \cite{Kingma_2014}, with initial learning rate 0.002, divided by 10 after the 5th and the 9th epoch. All ResNet's layers are frozen. Empirically, we observed meaningful results when $\lambda_f = 1.5$, $\lambda_P = 1$ and $\lambda_{OS} = 1$. All classification losses are Cross-Entropy losses.}. 
	We first compare our method against reported as well as re-implemented results of past approaches (Table~\ref{tab:Results}), then we further discuss our system's components (Table~\ref{tab:Ablation}). Our PyTorch code is publicly available\footnote{\url{https://bitbucket.org/deeplabai/vrd}. We train our model for 10 epochs using batch size 32 and Adam optimizer \cite{Kingma_2014}, with initial learning rate 0.002, divided by 10 after the 5th and the 9th epoch. ResNet-101 \cite{He_2015} is used for feature extraction and its layers are frozen. Empirically, we observe meaningful results when $\lambda_f = 1.5$, $\lambda_P = 1$ and $\lambda_{OS} = 1$. All classification losses are Cross-Entropy losses.}.
	
	\textit{Quantitative Results:}
	To address the ambiguity in the metric that prior works have used (Sec.~\ref{sec4.1}), we sort the methods based on $k$ and report results for all such choices. To the best of our knowledge, we are the first to review all past works that report Predicate Classification results on VRD, see Table~\ref{tab:Results}, where it is clear that our method outperforms prior works in all three metric settings, without the need of external corpora \cite{Yu_2017} or fine-tuning of the feature extractor \cite{Zhuang_2017}. Specifically, comparisons using $\mathbf{R_1@50}$ show that we achieve 0.43\%, 2.41\% and 4.76\% respective relative improvement over the main competitors \cite{Yin_2018,Yu_2017,Zhuang_2017}. Using $\mathbf{R_{70}@100}$, the main competitors are different \cite{Yin_2018,Cui_2018,Liang_2018} but still MATransE achieves 32\%, 40.1\% and 45.8\% relative error decrease.
	
	To extend the comparisons to the literature even further, we re-implement and fine-tune several approaches previously not comparable. In most cases, fine-tuning boosted the results over the originally reported. Moreover, we validate that the different metric settings do not always converge to the same solution. An interesting observation, for instance, is that although \cite{Dai_2017} largely outperforms \cite{Lu_2016} based on $\mathbf{R_1@50}$, their $\mathbf{R_{70}@100}$ performance is very similar. Even after improving the other methods' results, we still achieve better performance than all re-implemented ones.

	\begin{figure}[!t]
		\centering
		\includegraphics[width=.45\textwidth]{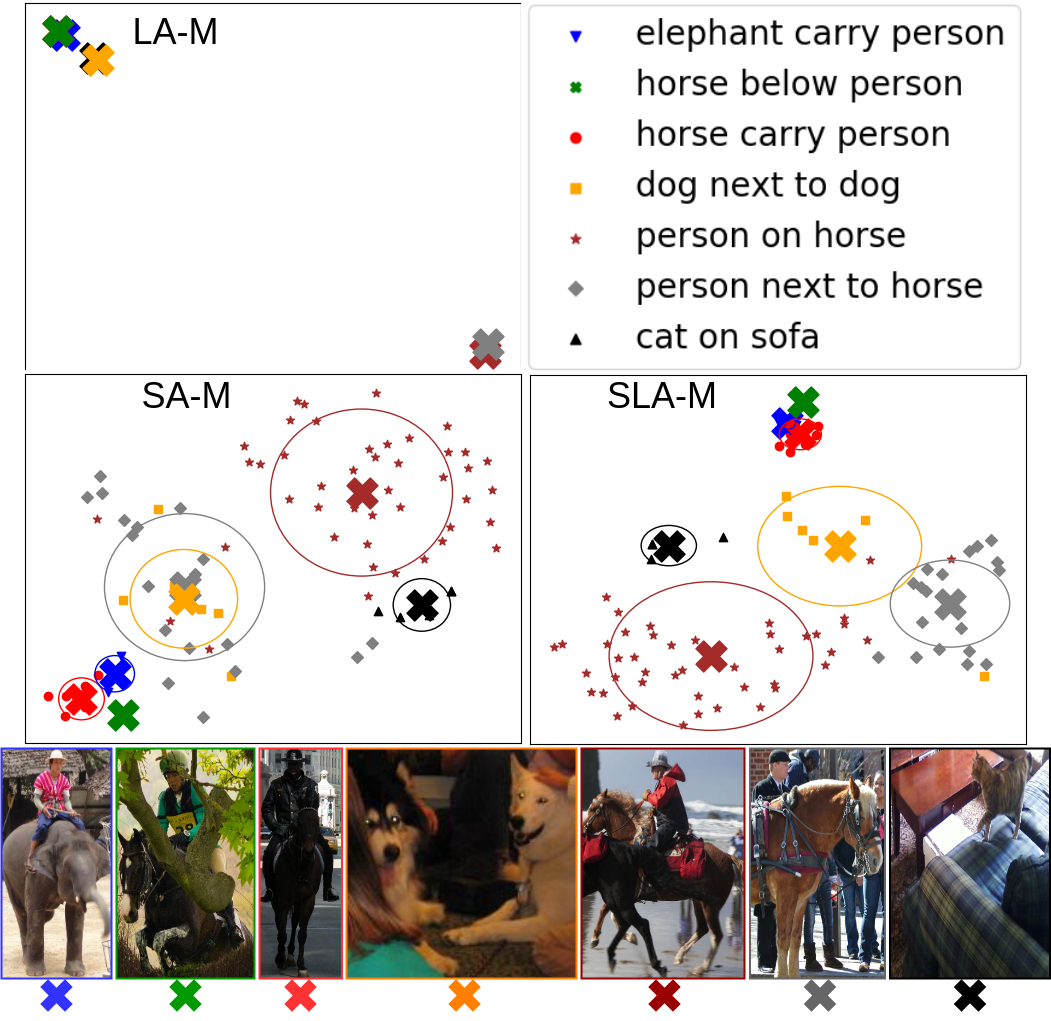}%[width=2.9in,height=1.6in]s
		\caption{Visualization \cite{Maaten_2008} of the projection space created by each attentional module; 7 classes. Centroids and std ellipsoids are also shown. LA-M quantizes relationships based exclusively on subject and object embeddings. SA-M generates weights that are highly-dependent on the objects’ relative positions. SLA-M exploits both spatial and semantic features to create a space where different relationships are better separated, while preserving the desired similarities. Best viewed in color.}
		\label{clustering}
	\end{figure}

	\begin{figure}[!t]
		\centering
		\includegraphics[width=.5\textwidth]{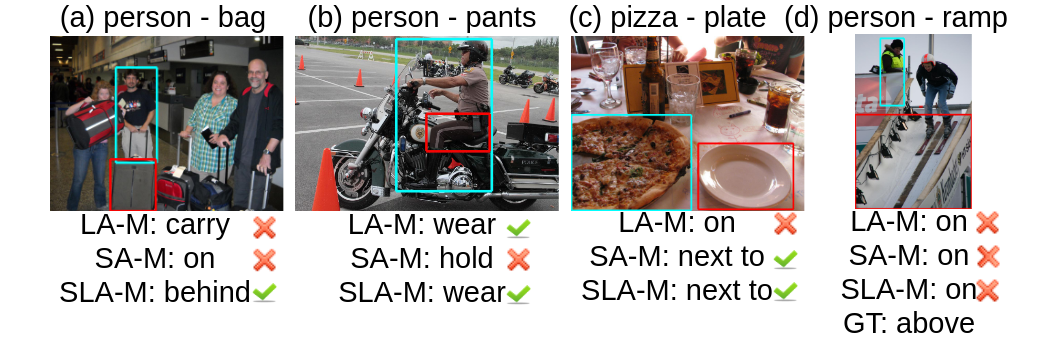}%[width=2.9in,height=1.6in]s
		\caption{Qualitative results using different attention modules are reported below the corresponding image. Ticks and crosses mark correct and incorrect results respectively. Subject and object boxes are in cyan and red respectively. (d) is a failure case (GT: ground-truth).}
		\label{qualitative}
	\end{figure}
	
	\begin{table}
		\centering
		\footnotesize
		\begin{tabular}{*{3}{c|}c}
			\hline
			\textbf{Method} & $\mathbf{R_{1}@50}$ & $\mathbf{R_{70}@50}$ & $\mathbf{R_{70}@100}$ \\ \hline
			
			(P + OS) + DS & 46.43 & 82.36 & 92.03 \\ \hline
			
			LA-M + (P + OS) + DS & 54.56 & 88.31 & 95.59 \\ \hline
			
			SA-M + (P + OS) + DS & 49.83 & 85.46 & 93.39 \\ \hline
			
			SLA-M + P & 55.82 & 89.6 & 95.97 \\ \hline
			
			SLA-M + OS & 55.85 & 89.72 & 96.22 \\ \hline
			
			SLA-M + (P + OS) & 55.14 & 88.29 & 95.06 \\ \hline
			
			\textbf{SLA-M} + \textbf{(P + OS)} + \textbf{DS} & \textbf{56.14} & \textbf{89.79} & \textbf{96.26} \\ \hline
		\end{tabular}
		\caption{Experimental comparisons by removing one component at a time. We get best results with the full system, that combines the Spatio-Linguistic Attention module (SLA-M) and P- and OS-branches, jointly trained with Deep Supervision (DS), outperforming the  linguistic (LA-M) or spatial attention (SA-M) or single branches.} 
		\label{tab:Ablation}
	\end{table}
	
%	\textbf{\textit{Ablation Study:}} Our system's strong points are the SLA-Module and the two-branch architecture aligned with Deep Supervision. We first test our network's performance by totally removing \textbf{SLA-M} or by replacing it with \textbf{LA-M} (pure linguistic) or \textbf{SA-M} (pure spatial) attention module. Next, we remove each one of the \textbf{P} and \textbf{OS} branch to test their standalone performance. Lastly, Deep Supervision's (\textbf{DS}) contribution is checked. Aware of the random procedures that take place during a network's training (e.g. weight initialization, batches' shuffling), we train our models 5 times and report mean performance. The results can be viewed in Table~\ref{tab:Ablation}.
	
	\textit{Spatio-Linguistic Attention:} We test our network's performance by totally removing SLA-M or by replacing it with LA-M (linguistic) or SA-M (spatial) attention module. All attentional modules clearly outperform the case without attentional mechanism, as they lead the network's ``focus" on the most discriminative features (Table~\ref{tab:Ablation}). Moreover, SLA-M clearly outperforms both LA-M, SA-M.
	
	\textit{Two-branch architecture and Deep Supervision:} Although competent, none of the single branches alone performs on par with the jointly trained 2-branch MATransE (Table~\ref{tab:Ablation}). This is not true when Deep Supervision is not used, as removing $\mathcal{L}_{P}$ and $\mathcal{L}_{OS}$ from Eq.~\ref{eqTotalLoss} cancels Eq.~\ref{eqMATransE}: we measure the mean value of $\norm{P - (O - S)}_2$ on the test set and find it equal to 0.187 when Deep Supervision is used and 0.722 otherwise. In the latter case, redundant parameters obstruct convergence to an optimal solution, proving the importance of MATransE's formulation.
	
	\textit{Qualitative Results:}
	Figure~\ref{clustering} is a visualization \cite{Maaten_2008} of the projection space created by each attentional module. Since LA-M exploits only subject and object embeddings, it quantizes the vector space and generates similar attentional weights for semantically similar object pairs, e.g. ``horse - next to - person" and ``elephant - carry - person". On the other hand, SA-M generates weights that are highly-dependent on the objects' positions and forms three ``super-clusters": ``carry - below", ``next to" and ``on". SLA-M combines the benefits of both modules and creates a space where the different relationships are separated, while preserving the desired neighborhoods. For instance, ``dog - next to dog" is far from ``person - next to - horse", a benefit of linguistic similarity, ``person - next to - horse" is far from ``person - on - horse", a benefit of spatial similarity and ``horse - below - person" is very close to ``elephant - carry - person", a benefit of both spatial and linguistic similarity. Note that in Fig.~\ref{clustering}, SA-M has projected ``dog - next to dog" very close to ``person - next to - horse", while this is resolved in SLA-M projections.
	
	We notice a similar behavior on Fig.~\ref{qualitative}, where the system that employs LA-M tends to predict the most semantically common relationships, e.g. it is more common for a pizza to be on a plate than next to it. SA-M favors predictions that share spatial configurations, e.g. in Fig.~\ref{qualitative}(b) the relative position of subject and object bounding boxes is such that SA-M favors ``hold". SLA-M is more robust than both SA-M and LA-M but may fail when both spatial and semantic information are ``distractive" (Fig.~\ref{qualitative}(d)).

%% file: tex_files/section_5.tex
We address the challenging task of Visual Relationship Detection introducing MATransE, a novel deeply supervised network that uses spatio-linguistic attention to drive the features of two branches and aligns their scores into a common space. Our experiments prove the significance of MATransE's components, by achieving the best scores on VRD dataset under all different metric settings. A future direction would be to test our method on the newly introduced and larger Visual Genome \cite{krishnavisualgenome}. Lastly, we show that our spatio-linguistic module creates a projection space where relationship triplets are represented as multimodal embeddings with desired properties. Their comparison with language models could be investigated to create knowledge representation priors using spatio-linguistic embeddings. 